# Health Monitoring of Movement Disorder Subject based on Diamond Stacked Sparse Autoencoder Ensemble Model

[#]Likun Tang[a]*, [#]Jie Ma[b], Yongming Li[b]*

([a]Chengdu Yangjun Technology Co., Ltd, Chengdu, P.R.China;

[b]School of Microelectronics and Communication Engineering, Chongqing University, Chongqing, 400044, P.R. China)

# as co-first author

* as co-corresponding author

**Abstract**

The health monitoring of chronic diseases is very important for people with movement disorders because of their limited mobility and long duration of chronic diseases. Machine learning-based processing of data collected from the human with movement disorders using wearable sensors is an effective method currently available for health monitoring. However, wearable sensor systems are difficult to obtain high-quality and large amounts of data, which cannot meet the requirement for diagnostic accuracy. Moreover, existing machine learning methods do not handle this problem well. Feature learning is key to machine learning. To solve this problem, a health monitoring of movement disorder subject based on diamond stacked sparse autoencoder ensemble model (DsaeEM) is proposed in this paper. This algorithm has two major components. First, feature expansion is designed using feature-embedded stacked sparse autoencoder (FSSAE). Second, a feature reduction mechanism is designed to remove the redundancy among the expanded features. This mechanism includes L1 regularized feature-reduction algorithm and the improved manifold dimensionality reduction algorithm. This paper refers to the combined feature expansion and feature reduction mechanism as the diamond-like feature learning mechanism. The method is experimentally verified with several state of art algorithms and on two datasets. The results show that the proposed algorithm has higher accuracy apparently. In conclusion, this study developed an effective and feasible feature-learning algorithm for the recognition of chronic diseases.

# 1 Introduction

The most common health problem for people with movement disorders is chronic diseases. Due to the limited mobility of movement disorders subject and the long duration of chronic diseases, monitoring of chronic diseases is important for people with movement disorders. The current lack of wearable sensor monitoring systems limits health monitoring of movement disorder subject and thus affects the timely treatment of early chronic diseases in movement disorders patients. Wearable sensors can be used to continuously collect physiological signals from patients in a fast and then the obtained data can be processed by machine learning algorithms [1]. The structure of a machine learning based chronic disease recognition system for people with movement disorder is shown in Fig.1. Firstly, physiological data from the movement disorders subject is collected by the wearable sensors. Then, it is stored in a chronic disease dataset by transmission. After that, machine learning algorithm is used to process and analyses the data to get the diagnosis results back to the patient or hospital. For example, optical sensors and pressure sensors are used to obtain vital signs continuously; then feature extraction and classification are conducted to obtain the classification of chronic disease prediction [2]. Fused data is collected by a group of sensors then the data is input to classifier model for early predicting disease [3]. Many pieces of wearable medical sensor system can realize data collection but cannot provide high-quality data and timely medical diagnoses, and the feature learning in machine learning has a greater impact on recognition results. Therefore, the improvement of feature learning in machine learning algorithms for chronic diseases is important. It is major motivation of this paper.

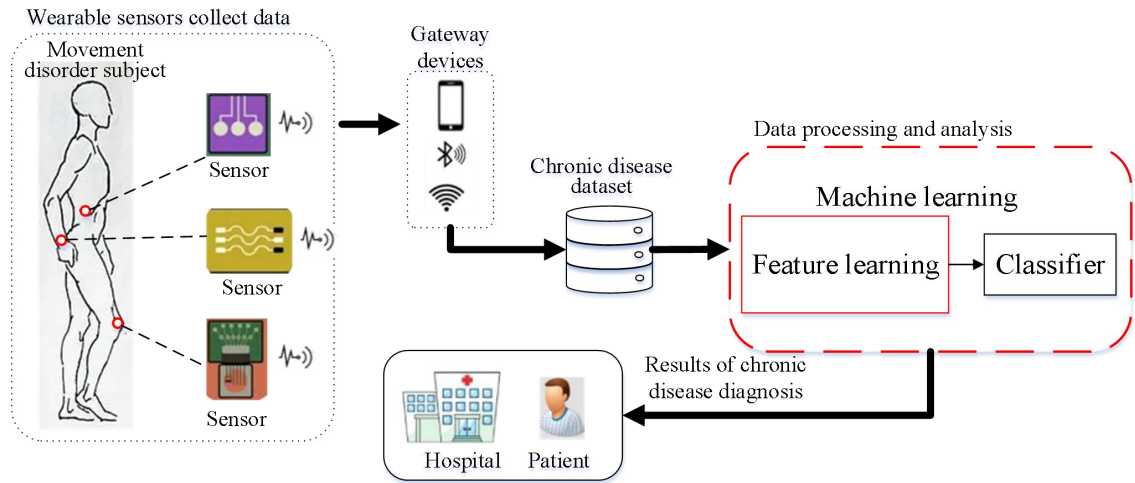

**Fig.1** The structure of a chronic diseases recognition system based on machine learning for people with movement disorder

Machine-learning technology has enabled new tools for recognition of chronic diseases because of its efficiency in data processing. Research on machine-learning methods for chronic disease recognition is focused on the following two areas: feature learning and classifier design. Among them, feature learning is particularly important. Feature learning methods mainly include feature selection and feature extraction methods. Feature selection is the process of selecting best features among all the features that are useful to discriminate classes. Ahmed H et al. [4] used univariate feature selection and Relief in selecting important features from the dataset. Shrivas et al. [5] proposed a union-based feature selection technique for predicting chronic kidney disease. Chormunge et al. [6] realized a new correlation and cluster-based feature selection technique to reduce the dimensionality issue in data mining tasks. Jayaraman and Sultana [7] have combined particle swarm optimization and gravitational cuckoo search algorithms for managing the features that exist in heart disease classification systems.

Feature extraction converts features from a high-dimensional space to a low-dimensional space. Some techniques used in chronic disease diagnosis are, linear discriminant analysis (LDA), and generalized discriminant analysis (GDA), and principal component analysis (PCA) and so on. Paul A K et al. [8] used weighted least squares to selected effective attributes, and Rasitha [9] used LDA to classify hypothyroid disease. Mohamed et al. [10] used PCA to reduce the dimensions of medical data of type 2 diabetes.

Classifiers can discover hidden patterns in existing human physiological data. Common classifiers [11-15] include the naive Bayesian (NB), radial basis function network (RBF Network), decision tree (DT), logistic regression (LR), k-nearest neighbor (KNN), functional tree (FT), logistic model tree (LMT), support vector machine (SVM), multi-Layer perceptron (MLP), random forest (RF), and recurrent neural network (RNN). Ekanayake et al. [16] selected high-quality features and used 11 machine-learning methods for model training: KNN, support vector classifier (SVC) with a linear kernel, LR, decision tree classifier, SVC with radial basis function (RBF) kernel, Gaussian NB, RF, a classical neural network, extra trees classifier, Adaboost classifier, and extreme gradient boosting (XGB) classifier. Ge et al. [17] proposed a new multi-label neural network prediction method for chronic diseases, combined with a neural network and multi-label learning technology. Polat [18] used the SVM, KNN, RF, and LDA methods to classify medical databases after attribute weighting. Cheruku et al [19] introduced a new hybrid decision support to a bat optimization algorithm and rough set theory-based system. In this hybrid system, redundant feature is effectively reduced by generating fuzzy rules. Alhassan and Zainon [20] presented a deep-belief network for heart disease diagnosis.

**Motivation:** These techniques developed by these studies improve the performance for chronic diseases detection. In particular, feature learning can provide high-quality features, thus reducing classifier complexity and improving recognition performance. However, the existing feature learning algorithms did not consider the characteristics of datasets on chronic diseases well, which typically are: 1) small or medium-sized features and samples, usually with no more than 35 features and no more than 1500 samples according to the relevant literature; 2) complex correlation between features and class label (disease status); and 3) a requirement for high recognition accuracy. Deep learning has powerful feature extraction capabilities, but cannot obtain sufficiently high quality of features in the condition of small samples. Existing non-parameter or low-parameter feature-reduction algorithms, such as PCA and LDA, can considerably reduce the number of features in the case of small samples. However, they use the original features and do not work well when the original features have low quality. Therefore, it is important and challenging to obtain high-quality feature set in the case of a small or medium-sized features and samples.

To solve the problem, this paper proposes a health monitoring of movement disorder subject based on diamond stacked sparse autoencoder ensemble model (DsaeEM) is proposed in this paper. This algorithm has two major components. First, feature expansion is designed using feature-embedded stacked sparse autoencoder (FSSAE). Second, a feature reduction mechanism is designed to remove the redundancy among the expanded features. This mechanism includes L1 regularized feature-reduction algorithm and the improved manifold dimensionality reduction algorithm. This paper refers to the combined feature expansion and feature reduction mechanism as the diamond-like feature learning mechanism.

The main contributions of this study are as follows.

(1) A diamond-like feature learning mechanism is proposed, realizing feature expansion and feature reduction with a diamond topology. Traditional feature learning algorithm has low adaptivity, and the deep learning algorithm is dependent on large number of samples. Both them cannot meet the requirement for high accuracy from chronic disease recognition with low samples. The proposed algorithm first expands features to enhance the representation capability and then reduces features to enhance the generalization capability, thereby reducing the requirement for large number of samples.

(2) Existing deep stack autoencoders learn poor quality features at small sample sizes and have limited feature complementary fusion performance due to insufficient complementarity. The designed FSSAE network introduces original features into the training process and structure of the network to improve the complementarity between high-quality features and original features, thus achieving high-quality deep features in small and medium sample sizes.

(3) Existing feature reduction methods do not adequately select and extract features. The proposed staged feature reduction mechanism first selects the expanded features based on L1 regularization, then reduces the dimensionality of the most important features base on manifold learning. It makes the features more compact without losing useful information.

The organization of the remaining paper organized in the following: The proposed method is mainly described in section 2. In section 3, experimental results are analyzed and reported. Finally, the major contributions and possible limitations are discussed in section 4.

## 2 The proposed method

In this section, we introduce the proposed algorithm and the hybrid synchronization mode for wearable sensors for people with movement disorders. Figure 2(a) shows the hybrid synchronization mode of the sensor groups. Figure 2(b) shows a hybrid synchronization mode of posture sensors designed for patient movement disorder monitoring (individual sensor groups are shown in the circles), which is easy to wear, low radiation and safe. Then, diamond stacked sparse autoencoder ensemble model (DsaeEM) is developed to better recognize chronic diseases according to the characteristics of chronic disease data. This algorithm consists of two parts, as illustrated in Fig. 3. First, feature expansion is designed using feature-embedded stacked sparse autoencoder (FSSAE). Second, a staged feature reduction mechanism is designed to remove the redundancy among the expanded features. It includes L1 regularization and weighted locality preserving discriminant projection.

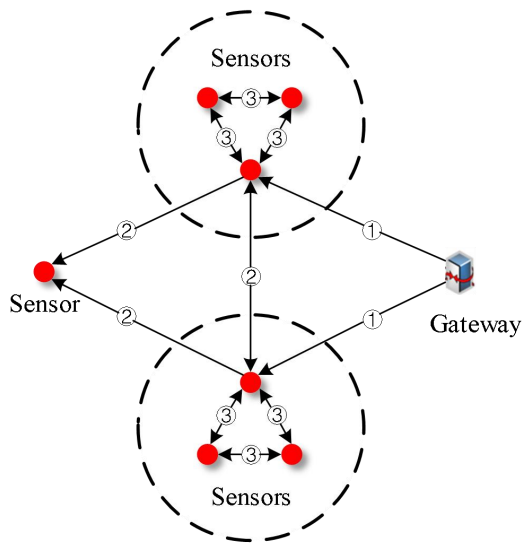 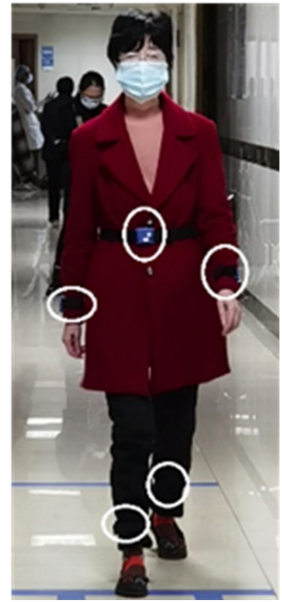

(a) Hybrid synchronous mode for sensor groups

(b) Schematic diagram of human motion awareness based on hybrid synchronous mode

**Fig.2** Hybrid synchronization mode for wearable sensors for people with movement disorders

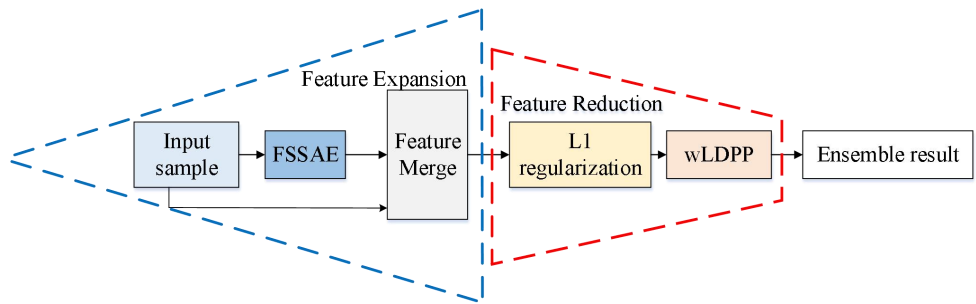

**Fig. 3** Proposed machine learning method

The main symbols used in this paper are listed in with their meanings in Table 1.

Table 1 Main symbols and their meanings

| Symbol | Meaning |
|---|---|
| $\mathbf{X}^{(o)} \in R^{N \times M}$ | Input data. N and M are the sample size and feature size. |
| $\mathbf{W}_1, \mathbf{W}_2$ | Weight matrixes of the encoder and decoder. |
| $\mathbf{b}_1, \mathbf{b}_2$ | Bias vector of the encoder and decoder. |
| $d^{(k)}$ | Number of hidden-layer units of the kth autoencoder. |
| $\lambda$ | Coefficient for the L2 weight regularization term. |
| $\beta$ | Coefficient for the sparsity regularization term. |
| $\rho$ | Sparse parameter. |
| $\hat{\rho}_j$ | Average activation value of all training samples on the $j$-th hidden neuron. |
| $S_B^{\phi}$ | Inter-class variance matrix. |
| $S_W^{\phi}$ | Intra-class variance matrix. |

## 2.1 Feature expansion mechanism

In the proposed method, the improved FSSAE, a lightweight deep network. The deep features are then combined with the original features for further feature expansion. FSSAE is improved compared to a stacked sparse autoencoder (SSAE). The FSSAE model is shown in Fig. 4.

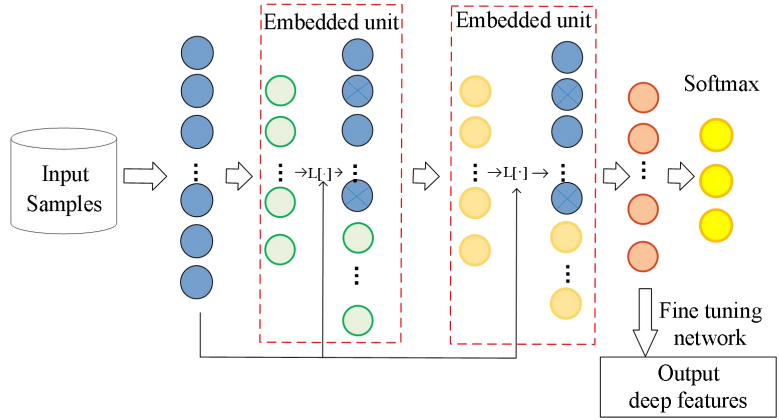

Fig. 4 FSSAE model

The key element of the FSSAE is the embedding combination between two adjacent autoencoder. Let $\mathbf{H}^{(k)} = [h_1^{(k)}, h_2^{(k)}, ..., h_N^{(k)}]$ denote the $k$th hidden layer's output matrix and $\mathbf{X}^{(o)} \in R^{N \times M}$ the sample input to the first layer (original input) of the FSSAE. Firstly, the output of the previous hidden layer $\mathbf{H}^{(k-1)}$ is concatenated with the original input $\mathbf{X}^{(o)}$ to obtain the combined feature as follows:

$$\mathbf{E}^{(k)} = [(\mathbf{X}^{(o)})^T; \mathbf{H}^{(k-1)}]. \tag{1}$$

Then, the combined feature $\mathbf{E}^{(k)}$ is transformed as follows:

$$L(\mathbf{E}^{(k)}) = \mathbf{G}^T \mathbf{E}^{(k)}, \tag{2}$$

where $\mathbf{G}$ is the appropriate sparse transformation matrix. Therefore, the objective function of the feature-embedded unit is defined as follows to filter partially redundant features:

$$\begin{aligned} \max_{\mathbf{G}} \quad & tr(\mathbf{G}^T \mathbf{E}^{(k)} (\mathbf{E}^{(k)})^T \mathbf{G}) \\ s.t. \quad & \sum G_{ij} = d \end{aligned} \tag{3}$$

Where $d$ is the number of high-quality features retained in feature extraction. After processing by the feature-embedded unit, the input data of the $k$th autoencoder are $L(\mathbf{E}^{(k)})$.

The output features of the hidden layer are divided into two groups, and the ratio of the two groups of features is kept consistent with the ratio of the two types of features in the input data of the encoder for that layer. The hidden-layer output features are expressed as $H = [H_{\Gamma_1}, H_{\Gamma_2}]$. The addition group sparsity constraint is represented as follows:

$$\psi(H) = \sum_{g=1}^{2} \| H_{\Gamma_g}^{(k)} \|_1 \tag{4}$$

After introducing embedded elements in the structure and sparse constraints in the training process, the objective function of the $k$-th autoencoder ($k > 1$) of the FSSAE is expressed as follows:

$$\begin{aligned} \arg\min_{\theta} \quad & \frac{1}{N} \sum_{i=1}^{N} \| L(\mathbf{E}^{(k)}) - L'(\mathbf{E}^{(k)}) \|^2 \\ & + \lambda(\| W_{k1} \|_2 + \| W_{k2} \|_2) + \beta(\sum_{j=1}^{d^{(k)}} KL(\rho \| \hat{\rho}_j) + \sum_{g=1}^{2} \| H_{\Gamma_g}^{(k)} \|_1) \end{aligned} \tag{5}$$

where $L'(\mathbf{E}^{(k)})$ is the $k$-th autoencoder's output. $\lambda$ and $\beta$ denote the regularization coefficient and sparsity coefficient, respectively. $\rho$ denotes the sparse parameter. The numerical value increases monotonously as difference between $\rho$ and $\hat{\rho}_j$ increases.

The proposed FSSAE algorithm is summarized in Algorithm 1.

**Algorithm 1:** Feature expansion algorithm
**Input:** Samples $\mathbf{X}^{(o)}$

1: Set parameters: $\lambda$, $\beta$, $\rho$, $d^{(k)}$, number of iterations.
2: **Pretraining:**
3:     Train the first layer of FSSAE and extract the output of the hidden layer $H^{(1)}$
4:     **For k = 2:K**
5:         Calculate transformation matrix G
6:         Calculate the output of the feature-embedded unit $L(\mathbf{E}^{(k)})$ by Eqs. (1–2)
7:         Train $k$th layer of FSSAE with objective function in Eq. 5
8:         Extract output $H^{(k)}$ of $k$-th hidden layer
9:     **End For**
10: **End Pretraining**
11: Stack hidden layers and add softmax layer
12: Fine-tune entire network based on back-propagation of scaled conjugate gradient algorithm
13: Output of last hidden layer are the learned deep features
**Output:** Deep features data $\hat{\mathbf{X}} \in R^{N*\hat{M}}$

## 2.2 Staged feature reduction mechanism

The feature-reduction mechanism consists L1 regularization and weighted locality preserving discriminant projection.

The L1 regularization, a common feature selection method, which obtains a sparse feature vector. The optimization objective function is as follows:

$$\arg\min_{\theta} \sum_{i=1}^{N}(y_i - \sum_{m=1}^{\hat{M}} \theta_m \hat{x}_{im})^2 + \mathfrak{A}\sum_{m=1}^{\hat{M}} |\theta_m|, \qquad (6)$$

where the first term is original error term and the second term is L1 regularization term. $\alpha$ denotes regularization factor. $\theta_m$ is $m$th feature's regression coefficient. Eq. 6 is optimized using the proximal gradient-descent method, and each gradient-descent iteration is

$$\theta^{(k+1)} = \arg\min_{\theta} \frac{A}{2} \|\theta - z\|_2^2 + \mathfrak{A}\|\theta\|_1, \qquad (7)$$

where $z = \theta^{(k)} - A^{-1}\nabla f(\theta^{(k)})$, and $f(\theta^{(k)}) = \sum_{i=1}^{N}(y_i - (\theta^{(k)})^T \hat{x}_i)^2$. $A$ is a constant greater than zero, and the components of $\theta = [\theta_1, \theta_2, ..., \theta_m, ..., \theta_{(\hat{M})}]$ are independent of each other. The solution of the L1 regular expression obtained using the iterative soft-threshold function can be expressed as follows:

$$\theta^{(k+1)} = soft_{\alpha A^{-1}}(\theta^{(k)} - A^{-1}\nabla f(\theta^{(k)})) = sign(\theta^{(k)})(|\theta^{(k)}| - \frac{\mathfrak{A}}{A}), \qquad (8)$$

where $sign(\cdot)$ is a symbolic function. The corresponding non-zero component features in $\theta^{(k+1)}$ are selected as the feature subset, that is, features selected by L1 regularization.

Then we will introduce proposed weighted locality-preserving discriminant projection (Wlpdp). $\tilde{\mathbf{X}}$ denotes the samples after the first stage of reduction, $C$ is the total number of classes. After sampling, the total number of samples is $N_s$, and the number of samples for the $c$th class is $N_{sc}$. The inter-class variance of the $N_s$ nearest-neighbor samples of the local sample center of the sample set $\tilde{\mathbf{X}}$ is $S_B^\phi$, defined as:

$$S_B^\phi = \sum_{c=1}^{C} \Delta_B^c (\Delta_B^c)^T , \qquad (9)$$

where $\Delta_B^c$ is the difference between the local sample center of the $c$th class and the local sample center, that is, $\Delta_B^c = \frac{1}{N_{sc}} \sum_{i=1,}^{N_{sc}} \tilde{x}_i^{(c)} - \frac{1}{N_s} \sum_{i=1}^{N_s} \tilde{x}_i$. The intra-class variance matrix of the $N_{sc}$ nearest-neighbor samples of the class center of the $c$th class sample is $S_W^\phi$, defined as

$$S_W^\phi = \sum_{c=1}^{C} \sum_{i=1}^{N_{Sc}} \Delta_i^c (\Delta_i^c)^T , \qquad (10)$$

where $\Delta_i^c$ is the difference between the $i$th sample of class $c$ and the local sample class center of class $c$, that is, $\Delta_i^c = \tilde{x}_i^{(c)} - \frac{1}{N_{sc}} \sum_{i=1,}^{N_{sc}} \tilde{x}_i^{(c)}$.

The regularization term is as follows:

$$\begin{aligned} J(Q) &= \sum_{i=1}^{N} \sum_{j=1}^{N} \left\| Q^T \tilde{x}_i - Q^T \tilde{x}_j \right\|^2 W_{ij} \\ &= Tr(\sum_{i=1}^{N} \sum_{j=1}^{N} (2 Q^T \tilde{x}_i \tilde{x}_i^T Q - 2 Q^T \tilde{x}_i \tilde{x}_j^T Q) W_{ij}) , \qquad (11) \\ &= Tr(\sum_{i=1}^{N} Q^T \tilde{x}_i D_{ii} \tilde{x}_i^T Q - \sum_{i=1}^{N} \sum_{j=1}^{N} W_{ij} Q^T \tilde{x}_i \tilde{x}_j^T Q) \end{aligned}$$

where $D_{ii} = \sum_{j=1}^{N} W_{ij}$ is the diagonal matrix. $W$ is the affinity matrix. The matrix form of Eq. 11 can be written as $Tr(Q^T \tilde{X}(D-W)\tilde{X}^T Q)$. Setting $L = D - W$, the locality preservation regularization term can be expressed as

$$J(Q) = Tr(Q^T \tilde{X} L \tilde{X}^T Q) . \qquad (12)$$

Using Eqs. (9–12), the proposed Wlpdp can be expressed as:

$$\min_{Q} \quad Tr(Q^T S_W^\phi Q)$$
$$s.t. \quad Tr(Q^T S_B^\phi Q) - \gamma J(Q) = \kappa I \quad , \tag{13}$$

where $\kappa$ is a constant and $\gamma$ represents the Lagrange penalty factor. Through adding the Lagrange multiplier $\eta$, the Eq. 13 rewritten as

$$L(Q,\eta) = Tr(Q^T S_W^\phi Q) - \eta(Q^T S_B^\phi Q - \gamma J(Q) - \kappa I) \quad . \tag{14}$$

Taking the partial derivative of Eq 14 and setting $\dfrac{\partial L(Q,\eta)}{\partial Q} = 0$, the result is

$$\frac{S_W^\phi Q}{(S_B^\phi - \gamma \tilde{X} L \tilde{X}^T)} = \eta Q \quad . \tag{15}$$

After obtaining the projection matrix $Q$, we take the top $l$ eigenvectors corresponding to the biggest eigenvalues of $Q$ to obtain projection matrix $Q_l$. Using the locality-preserving discriminant projection in each subspace separately, we can obtain $T$ projection matrices: $Q_l^1$, $Q_l^2$ …, $Q_l^T$. The final projection matrix $\mathbf{Q}_l^E$ is obtained through the weighted integration of $Q_l^t$ in each subspace projection transformation.

In addition, an ensemble-learning method is adopted for the fusion mechanism. Specifically, let the sampling ratios of samples and features be $\delta_S$ and $\delta_F$, respectively, and the mixed-feature dataset be sampled according to the bagging strategy $r$ times, forming $r$ subsets. The proposed staged feature-reduction mechanism is then summarized in Algorithm 2.

---

**Algorithm 2:** Staged feature reduction mechanism
**Input:** Data $\hat{X}$ after feature expansion
**1: First-stage feature reduction**
2:    Hybrid feature selection for $\hat{X}$ based on Eqs. (6–8)
**3: End first-stage feature reduction**
**4: Second-stage feature reduction**
5:    $\delta_S = 0.7, \delta_F = 0.5$.
6:    Sample $t$ times to form $t$ subsets.
7:    Train rth SVM:
8:    For i = 1:T
9:        Choose $n_s$ train samples randomly
10:       Calculate scatter matrixes $S_B^\phi$ and $S_W^\phi$ using Eqs. (9–10)
11:       Calculate diagonal matrix **D** and the Laplacian matrix **L**

| | |
|---|---|
| **12:** | Solve mapping matrix **Q** using Eq. 15 |
| **13:** | **End For** |
| **14:** | Search for optimal weight to obtain final mapping matrix $\mathbf{Q}_l^E$ |
| **15:** | Map rth subset to train SVM. |
| **16:** | **End second-stage feature reduction** |
| **17:** | Obtain the ultimate class label |
| **Output:** | Predicted label |

# 3 Experimental results and analysis

Two sets of experiments are conducted to verify the effectiveness of the algorithm in this paper. The first experiment compares the proposed algorithm with existing representative feature learning algorithms, representative autoencoders and the state-of-the-art algorithms to verify the proposed methods in this paper. The second experiment analyses the effects of some important parameters including coefficient for the L2 weight regularization, sparsity regularization coefficient and sparse parameter.

## 3.1 Experimental conditions

The performance of the proposed algorithm is tested on several relevant used datasets. Two publicly available chronic disease datasets (Pima Indians Diabetes and Statlog Heart Data Set) are selected for the experiments; they include data on cardiovascular diseases and diabetes, which are the two major chronic diseases [21–22]. The basic information on the datasets is presented in Table 2. The Pima Indians Diabetes and Statlog Heart Data Set are representative datasets for diabetes and heart disease, respectively.

Table 2 Basic information of datasets used in study

| Dataset | Instances | Attributes | Class | Relevant paper |
|---|---|---|---|---|
| Statlog Heart Data Set (Heart) | 270 | 13 | 2 | Reference [23] |
| Pima Indians Diabetes Data Set (PID) | 768 | 8 | 2 | Reference [24] |

We set three AEs in the proposed FSSAE. The number of hidden units (neurons in hidden layer) is selected by considering the dimensional range of the feature vectors for different datasets, and a grid search is used to find the best values. The adjustable parameters in the proposed FSSAE include coefficient for regularization and sparsity parameters. The relevant parameters are listed in Table 3, including three parameters of tuned FSSAE objective function and the number of neural units of the FSSAE hidden layer for each dataset.

Table 3 Parameter information

| Parameter | Parameter meaning | Parameter value |
|---|---|---|
| $\lambda$ | Coefficient for the L2 weight regularization term. | 10e−5, 10e−4, 10e−3 |
| $\beta$ | Coefficient for the sparsity regularization term. | 1,2,3,4,5,6 |
| $\rho$ | Sparse parameter. | [0.02, 0.1] |
| PID hidden units | Number of FSSAE hidden-layer units in PID dataset | 120–40–16 |
| Heart hidden units | Number of FSSAE hidden-layer units in Heart dataset | 100–60–24 |

The K-fold cross validation method is used in the experiments. A unified experimental environment is used: Windows 10 64-bit operating system with 128 GB memory. The programming tool is R2018b MATLAB.

## 3.2 Evaluation criteria

Accuracy (Acc), sensitivity (Sens), precision (Prec), F1_score, and specificity (Spec), which are calculated from the confusion matrix, are chosen as metrics to assess the effectiveness of the model. The confusion matrix or error matrix is a useful tool used to visualize the overall performance of classifiers. The datasets used in this study are multi-classified and binary, with a 2 × 2 confusion matrix for binary and n × n confusion matrix for multi-classification. The Acc, Sen, Pre, Spec, and F1_score of each category can be expressed as follows.

$$Acc = \frac{TP + TN}{TP + FP + FN + TN} \quad (16)$$

$$Prec = \frac{TP}{TP + FP} \quad (17)$$

$$Sens = \frac{TP}{TP + FN} \quad (18)$$

$$Spec = \frac{TN}{TN + FP} \quad (19)$$

$$F1\_score = \frac{2(Prec \times Sens)}{Prec + Sens} \quad (20)$$

## 3.3 Algorithm comparison
### 3.3.1 Comparison with typical feature-learning algorithms

To evaluate the performance of the DsaeEM algorithm, its performance is compared to those of feature-learning methods that are representative, including feature selection Least absolute shrinkage and selection operator (LASSO) [25] and Relief [26]) and feature extraction algorithms (LDA [27], and locality preserving projection (LPP) [28]). The SVM is used as the base classifier, since it is a commonly used classifier. The results are listed in Table 4.

Table 4 Comparison of different feature-learning algorithms

| Dataset | Performance indices | LDA (%) | LPP (%) | Relief (%) | LASSO (%) | Proposed method (%) |
|---|---|---|---|---|---|---|
| PID | Acc | 80.21 | 80.03 | 78.17 | 77.59 | **84.54** |
| | Prec | 80.09 | 79.38 | 78.30 | 77.76 | **88.50** |
| | Sens | 80.58 | 81.33 | 78.34 | 77.59 | **80.99** |
| | Spec | 79.85 | 78.74 | 77.99 | 77.60 | **88.11** |
| | F1_score | 80.30 | 80.25 | 78.24 | 77.67 | **84.18** |
| Heart | Acc | 94.17 | 94.58 | 93.33 | 88.75 | **96.67** |
| | Prec | 95.12 | 94.36 | 94.85 | 89.54 | **97.59** |
| | Sens | 93.33 | 95.00 | 91.67 | 88.33 | **95.83** |
| | Spec | 95.00 | 94.17 | 95.00 | 89.17 | **97.50** |
| | F1_score | 94.12 | 94.61 | 93.19 | 88.76 | **96.65** |

The results in Table 4 show that the DsaeEM algorithm compared with the other feature-learning algorithms has superior performance algorithm can obtain better quality features than other feature learning algorithms, and can considerably improve the classification accuracy. The first possible reason is that the compared methods are based on the original features, but the proposed algorithm expands the original features, thereby obtaining higher-quality features efficiently. The second possible reason is the high effectiveness of the feature-reduction mechanism.

### 3.3.2 Comparison with recent chronic disease detection algorithms

To further verify the accuracy of DsaeEM algorithm, its performance is compared to those of state-of-the-art algorithms proposed in recent years for classifying chronic diseases, including those by Hasan *et al.* [29] and Guia *et al.* [30]. Table 5 shows that all the models perform better in either positive or negative chronic disease prediction, the two cross-validation methods (CV, including k-fold and Holdout) are used for fair comparison. Seen from the table, the proposed method outperforms the others in terms of either precision or F1_score or both. The cases are similar under 5-fold and Hold out CV.

Table 5 Performance comparison of proposed method with state-of-the-art chronic disease detection algorithms

| Dataset | Performance indices | Literature [29] | Literature [30] | Proposed method | Proposed method |
|---|---|---|---|---|---|

|  |  | (5-flod) | (Holdout) | (5-flod) | (Holdout) |
| --- | --- | --- | --- | --- | --- |
|  | Acc(%) | **88.8** | 80.04 | 84.54 | 83.18 |
|  | Sens(%) | 78.9 | 73.51 | **80.99** | 74.07 |
| PID | Spec (%) | **93.4** | - | 88.11 | 92.45 |
|  | Prec(%) | 84.2 | 84.93 | 88.50 | **90.91** |
|  | F1_score(%) | - | 78.80 | **84.18** | 81.63 |
|  | Acc(%) | 73.9 | 85.42 | **96.67** | 95.83 |
|  | Sens(%) | 74.8 | 75.00 | 95.83 | **100** |
| Heart | Spec (%) | 73.0 | - | **97.50** | 91.67 |
|  | Prec(%) | 69.7 | 94.74 | **97.59** | 92.31 |
|  | F1_score(%) | - | 83.72 | **96.65** | 96.00 |

## 3.4 Analysis of FSSAE parameters

The FSSAE model is a critical part of the DsaeEM. Therefore, it is important to have analysed the effect of the FSSAE parameters on the performance. First, the effect of the sparsity parameter on FSSAE is explored. In the experiments of this paper, it is chosen to be in the range of 0.02-0.1.

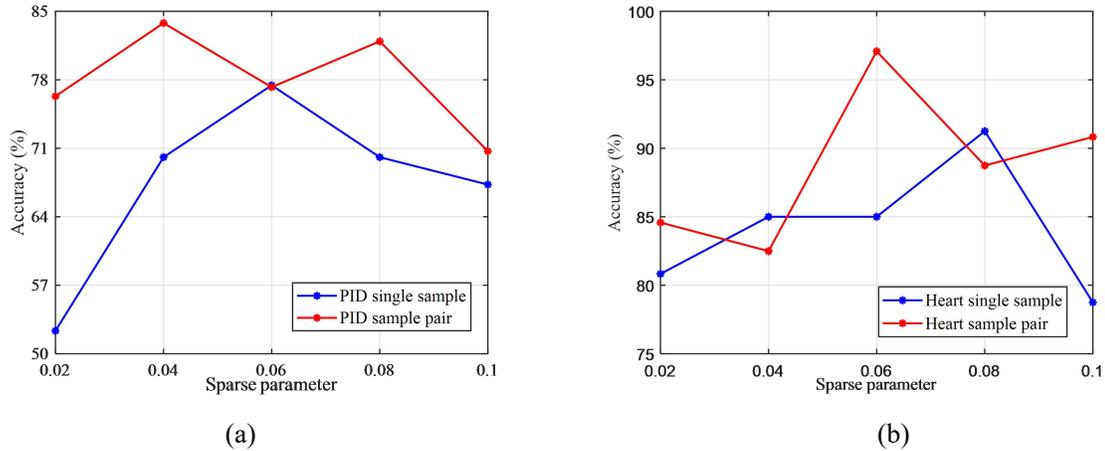

**Fig. 5** Effect of sparse proportion on FSSAE with (a) PID; (b) Heart datasets

As seen from Fig. 5, different sparse parameters give different results. The sparse parameter had an apparent impact on the accuracy of the FSSAE algorithm. The optimal sparse parameter is different for different datasets, and there is no fixed selection criterion. The datasets used have different numbers of samples, feature dimensions, and numbers of categories, and the optimal sparse parameter may be related to these factors.

The effects of the $\lambda$ in the range of $10^{-5}$ to $10^{-3}$ and $\beta$ in the range of 1 to 6 on the performance of the proposed FSSAE are analyzed together. Fig 6 shows that when the $\lambda$ is fixed, the β had relatively little effect on the network. When the β is fixed, the closer the λ is to 0, the more robust is the model.

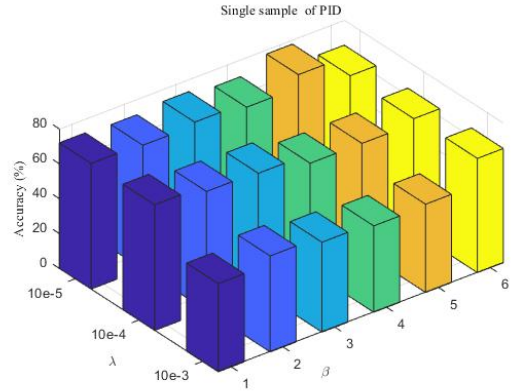

(a)

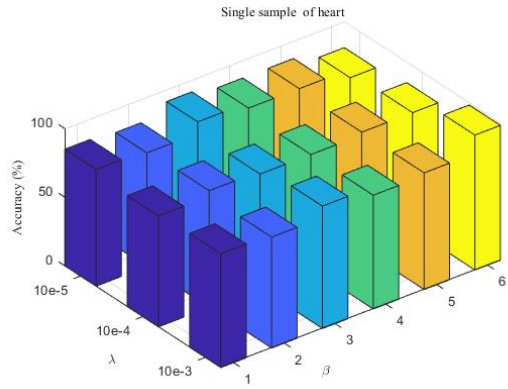

(b)

**Fig. 6** Effect of penalty term coefficients on FSSAE performance: (a) sample of PID; (b) sample of Heart

## 4 Discussion and conclusions

The detection of chronic diseases is very important for people with movement disorders because of their limited mobility and long duration of chronic diseases. Currently, a practical approach is to first use wearable sensors collect data from the human with movement disorders, and then process the data using machine learning. Machine learning algorithm is an effective tool for the analysis of sensing data of chronic disease. Therefore, it is important and challenging to study high efficient feature learning methods for sensor monitoring of chronic disease. Traditional feature learning methods are restricted from original features and cannot construct high-quality features, whereas deep feature learning methods suffer from the small-sample problems.

To overcome these limitations, this paper proposed a solution- a health monitoring of movement disorder subject based on diamond stacked sparse autoencoder ensemble model (DsaeEM). First, feature expansion is designed using feature-embedded stacked sparse autoencoder (FSSAE). Second, a staged feature reduction mechanism is designed to remove the redundancy among the expanded features. This mechanism includes L1 regularized feature-reduction algorithm and the improved manifold dimensionality reduction algorithm.

In the experimental section, two sets of experiments are organized to validate the effectiveness of the algorithm and to research the influence of algorithm's parameters. The results show that the proposed algorithm is effective apparently. Compared with feature extraction algorithms, such as LDA, the accuracy of DsaeEM's classification recognition is higher because the chronic disease samples are often random. However, LDA is not sensitive to non-Gaussian distributed samples. Compared with feature selection algorithms including Relief, the advantage of DsaeEM lies in its stronger learning ability and lower fault tolerance. Relief gives higher weights to all features that are highly relevant to the class, so the limitation of Relief is in that it does not effectively remove redundant features.

The proposed method can produce different variations, by using different feature learning algorithms and different classifiers. Therefore, the proposed method has good generalization. Compared with other deep learning algorithms, the autoencoder has lightweight structure and parameters. Therefore, it is discussed in this paper as representative deep neural network. That is major reason why only autoencoder rather than other deep learning methods is involved in this paper.

In future work, other types of deep neural networks can be considered for further verification and improvement. In addition, the proposed algorithm can be validated on more datasets and embedded in portable systems for practical diagnosis of chronic diseases in people with movement disorders.

## Acknowledgements

This work was also supported by Sichuan Science and Technology Fund (No. 21ZDYF3646).

## Conflict of interest

The authors declare no conflicts of interest pertaining to this work.

## Ethics Approval and Consent to Participate

Not applicable.

## Consent for publication

Not applicable.

## References

[1] Yin H, Jha N K. A health decision support system for disease diagnosis based on wearable medical sensors and machine learning ensembles[J]. IEEE Transactions on Multi-Scale Computing Systems, 2017, 3(4): 228-241.

[2] Wu J, Chang L, Yu G. Effective data decision-making and transmission system based on mobile health for chronic disease management in the elderly[J]. IEEE Systems Journal, 2020,15(4): 5537-5548.

[3] Muzammal M, Talat R, Sodhro A H, et al. A multi-sensor data fusion enabled ensemble approach for medical data from body sensor networks[J]. Information Fusion, 2020, 53: 155-164.

[4] Ahmed H, Younis E M G, Hendawi A, et al. Heart disease identification from patients' social posts, machine learning solution on spark[J]. Future Generation Computer Systems, 2020, 111: 714-722.

[5] Shrivas A K, Sahu S K, Hota H S. Classification of chronic kidney disease with proposed union based feature selection technique[J]. Social Science Research Network Electronic Journal, 2018, 4:26-27.

[6] Chormunge S, Jena S. Correlation based feature selection with clustering for high dimensional data[J]. Journal of Electrical Systems and Information Technology, 2018, 5(3): 542–549.

[7] Jayaraman V, Sultana H P. Artificial gravitational cuckoo search algorithm along with

particle bee optimized associative memory neural network for feature selection in heart disease classification[J]. Journal of Ambient Intelligence and Humanized Computing, 2019, 1–10.

[8] Paul A K, Shill P C, Rabin M, et al. Adaptive weighted fuzzy rule-based system for the risk level assessment of heart disease [J]. Applied Intelligence, 2018, 48(7): 1739-1756.

[9] Rasitha G. Predicting thyroid disease using linear discriminant analysis (LDA) data mining technique[J]. International Journal of Modern Trends in Engineering and Research, 2016, 4(1):4–6.

[10] Mohamed E I, Linder R, Perriello G, et al. Predicting type 2 diabetes using an electronic nose-based artificial neural network analysis[J]. Diabetes Nutrition and Metabolism, 2002, 15(4):215-221.

[11] Abreu P H, Santos M S, Abreu M H, et al. Predicting breast cancer recurrence using machine learning techniques: a systematic review[J]. ACM Computing Surveys, 2016, 49(3): 1-40.

[12] Yildirim, Pinar. Chronic kidney disease prediction on imbalanced data by multilayer perceptron: chronic kidney disease prediction[C]// 41st IEEE Computer Software and Applications Conference. IEEE, 2017:193-198.

[13] Zou Q, Qu K, Luo Y, et al. Predicting diabetes mellitus with machine learning techniques[J]. Frontiers in Genetics, 2018: 515.

[14] Rubini L J, Eswaran D P. Generating comparative analysis of early stage prediction of chronic kidney disease[J]. International Journal of Modern Engineering Research, 2015, 5(7): 49-55.

[15] Parul Sharma and Poonam Sinha. Comparative study of chronic kidney disease prediction using KNN and SVM[J]. International Journal of Engineering Research and Technology, 2015, 4(12): 608-612.

[16] Ekanayake I U, Herath D. Chronic kidney disease prediction using machine learning methods[C]// 2020 Moratuwa Engineering Research Conference. IEEE, 2020: 260-265.

[17] Ge R, Zhang R, Wu Q, et al. Prediction of chronic diseases with multi-label neural network[J]. IEEE Access, 2020, 127: 24-25.

[18] Polat, Kemal. Similarity-based attribute weighting methods via clustering algorithms in the classification of imbalanced medical datasets[J]. Neural Computing and Applications, 2018, 30(3): 987–1013.

[19] Cheruku R, Edla D R, Kuppili V, et al. RST-BatMiner: A fuzzy rule miner integrating rough set feature selection and bat optimization for detection of diabetes disease[J]. Applied Soft Computing, 2017, 67:764-780.

[20] Alhassan A M, Wan Zainon W M N. Taylor bird swarm algorithm based on deep belief network for heart disease diagnosis[J]. Applied Sciences, 2020, 10(18):6626.

[21] Hegde S, Mundada M R. Early prediction of chronic disease using an efficient machine learning algorithm through adaptive probabilistic divergence based feature selection approach[J]. International Journal of Pervasive Computing and Communications, 2020, 17(1): 20-36.

[22] Simon D S, Fraser, Paul J, et al. Kidney disease in the global burden of disease study 2017

[J]. Nature Reviews Nephrology, 2019, 15(4):193.

[23] Lichman M. UCI machine learning repository[J]. http://archive.ics.uci.edu/ml ,2013.

[24] Smith J W, Everhart J E, Dickson W C, et al. Using the ADAP learning algorithm to forecast the onset of diabetes mellitus[C]// Proceedings Symposium on Computer Applications in Medical Care, 1988, 10:261–265.

[25] Yamada M, Jitkrittum W, Sigal L, et al. High-dimensional feature selection by feature-wise kernelized lasso[J]. Neural Computation, 2014, 26(1): 185-207.

[26] Sun Y, Lou X, Bao B. A novel relief feature selection algorithm based on mean-variance model[J]. Journal of Information and Computational Science, 2011, 8(16): 3921-3929.

[27] Li C H, Kuo B C, Lin C T. LDA-based clustering algorithm and its application to an unsupervised feature extraction[J]. IEEE Transactions on Fuzzy systems, 2010, 19(1): 152-163.

[28] He X, Niyogi P. Locality preserving projections[J]. Advances in Neural Information Processing Systems, 2003, 16: 153-160.

[29] Hasan M K, Alam M A, Das D, et al. Diabetes prediction using ensembling of different machine learning classifiers[J]. IEEE Access, 2020, 8:76516-76531.

[30] Guia J, Concepcion R S, Bandala A A, et al. Performance comparison of classification algorithms for diagnosing chronic kidney disease[C]// 2019 IEEE 11th International Conference on Humanoid, Nanotechnology, Information Technology, Communication and Control, Environment, and Management. IEEE, 2019:1-7.